\documentclass[conference]{IEEEtran}
\IEEEoverridecommandlockouts

\usepackage{cite}
\usepackage{amsmath,amssymb,amsfonts}
\usepackage{algorithmic}
\usepackage{graphicx}
\usepackage{textcomp}
\usepackage{xcolor}
\usepackage{booktabs} 
\def\BibTeX{{\rm B\kern-.05em{\sc i\kern-.025em b}\kern-.08em
    T\kern-.1667em\lower.7ex\hbox{E}\kern-.125emX}}
\begin{document}

\title{EmbodiedUS-FS: Fast-Slow Intelligence for Ultrasound Robotics
}

\author{ Fangzhuo Zhang, Xinyu Wang, Xiao Yang, Jinchang Zhang, 
\thanks{Fangzhuo Zhang is with independent researcher, Xinyu Wang is with Shanghai World Foreign Language Academy,  Xiao Yang is with University of Wyoming, Jinchang Zhang is with SUNY Binghamton,USA
        {\tt\small chang2021cv@gmail.com}}%
}

\maketitle

\begin{abstract}
Robotic ultrasound scanning in real clinical environments requires both high-level semantic planning and low-level closed-loop execution: physicians’ natural-language instructions often contain implicit procedural logic, while execution is affected by patient motion, contact variations, and target drift. We propose a fast–slow hierarchical embodied ultrasound system. The Slow Brain performs intent parsing and stage-wise task planning with knowledge augmentation from an API/handbook corpus, and generates executable plans via task-graph construction and plan verification. The Fast Brain fuses multimodal feedback to refine local actions and perform recovery behaviors. The system further integrates a Safety Shield and a hierarchical escalation policy to constrain risks and trigger replanning or human confirmation in failure cases. Experiments on planning evaluation, closed-loop execution under dynamic perturbations, and safety-mechanism validation demonstrate that our hierarchical design significantly improves task success rates while reducing safety violations.
\end{abstract}


\section{Introduction}
Ultrasound imaging is a core technology in non-invasive diagnosis and is widely used in obstetrics, cardiovascular assessment, abdominal examination, and other clinical scenarios \cite{chan2010basics}, where it plays an important role in early screening, bedside assessment, and diagnostic decision-making \cite{mayo2019thoracic}. Recent advances in tactile sensing \cite{cao2023ultra}, compliant force control, robotic trajectory planning \cite{wang2022full}, and medical image analysis have enabled automated robotic ultrasound scanning. Meanwhile, recent studies on vision-language embodiment and interpretable medical robotics have shown the potential of intelligent robotic systems \cite{zhang2025vision,li2025automated}. However, deploying autonomous ultrasound robots in real clinical environments remains challenging because ultrasound acquisition is not merely a low-level control problem, but a knowledge-intensive clinical workflow that requires correct interpretation of physician instructions, adherence to examination protocols, real-time image-quality assessment, and safety-aware interaction with patients \cite{wang2022full}.

A key limitation of existing methods is that clinical instruction understanding, workflow-level reasoning, and execution-level closed-loop control are often coupled within a single decision-making pipeline, leading to a mismatch between clinical reasoning and real-time robotic execution \cite{knepper2015recovering}. Physicians’ spoken instructions usually contain implicit anatomical targets, procedural dependencies, quality requirements, and safety constraints, requiring the system to reason over clinical and operational knowledge in a globally consistent manner. In contrast, ultrasound scanning itself is highly dynamic and depends on real-time feedback from ultrasound images, robot states, contact forces, and patient motion. Patient micro-motion, contact variation, target drift, and image-quality fluctuation require high-frequency local adaptation. These two requirements differ substantially in temporal scale and decision objective: clinical workflow reasoning emphasizes semantic correctness, procedural validity, and safety compliance, whereas robotic execution emphasizes rapid feedback-driven adjustment and stable image acquisition. Satisfying both requirements within a single module is therefore difficult \cite{merouche2015robotic}.

To address this issue, we propose a fast--slow hierarchical embodied intelligence system for robotic ultrasound. The Slow Brain performs knowledge-grounded clinical workflow reasoning by leveraging ultrasound examination knowledge, robot API knowledge, and handbook-based procedural constraints for intent parsing, task decomposition, stage-wise planning, and plan verification. The Fast Brain performs high-frequency closed-loop action refinement and local recovery based on multimodal feedback, including real-time ultrasound images, robot pose/force states, and patient-motion information, thereby improving scan stability and image quality under dynamic clinical conditions. Furthermore, we introduce structured plan verification, a Safety Shield, and a hierarchical escalation mechanism to trigger Slow-Brain replanning or human confirmation when local failures persist, uncertainty increases, or safety bounds are violated. This design enables safe, interpretable, and controllable clinical ultrasound assistance by integrating healthcare knowledge reasoning with real-time embodied execution.
degraded execution.

\begin{figure*}[t]
\begin{center}
\includegraphics[width=17cm, height=7cm]{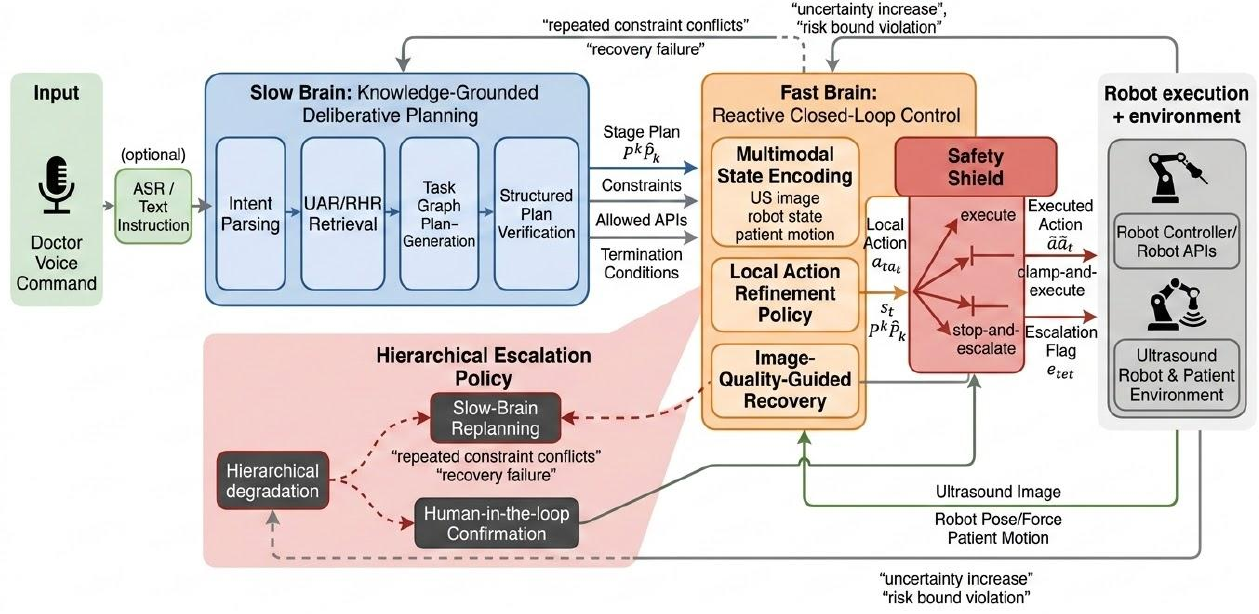}
\end{center}
\vspace{-5 mm}
\caption{Fast-slow hierarchical embodied ultrasound agent for robotic ultrasound scanning.
The Slow Brain performs knowledge-grounded stage planning from doctor voice instructions, while the Fast Brain performs reactive closed-loop control using multimodal feedback (ultrasound image, robot pose/force, and patient motion). A Safety Shield enforces execution constraints and a hierarchical escalation policy triggers Slow-Brain replanning or human confirmation under failures or risk-bound violations.}
\vspace{-6mm}
\label{arch}
\end{figure*}

Overall, our contributions are summarized as follows:
1. We propose a fast--slow hierarchical embodied intelligence framework for robotic ultrasound, which decouples knowledge-grounded clinical workflow planning from feedback-driven closed-loop execution to improve instruction understanding and dynamic scanning robustness.
2. We design a knowledge-enhanced planning and execution mechanism that integrates API/handbook retrieval, stage-wise task graph generation, structured plan verification, and image-quality-guided local recovery for interpretable and stable ultrasound task execution.
3. We introduce a safety-aware escalation strategy with a Safety Shield, which constrains risky actions and triggers Slow-Brain replanning or human confirmation under persistent failures, uncertainty, or safety-bound violations.
Our framework is shown in Fig. \ref{arch}.

\section{Relate Work}
\subsection{ Surgical Ultrasound Robotics}
Robotic ultrasound has attracted increasing attention in surgical and interventional settings, where intraoperative conditions often involve tissue deformation, patient/organ motion, and strict requirements for stable contact and consistent imaging. \cite{krupa2007real} Early studies established foundational paradigms of image-guided control and visual servoing, adjusting probe pose based on real-time ultrasound feedback, \cite{abolmaesumi2002image} and later advanced ultrasound visual servoing using intensity cues or image-moment features to improve robustness and controllability under challenging imaging conditions. \cite{nadeau2011intensity} Force-aware control further reduces the risk of excessive tissue compression, mitigates contact loss, and improves scanning stability by regulating probe–tissue interaction forces. \cite{virga2016automatic} Robotic platforms have also been proposed for continuous or repeatable intraoperative ultrasound acquisition to support real-time imaging updates during surgery. \cite{gumprecht2011robotics} For interventional workflows, robotic ultrasound has been integrated with needle insertion/navigation systems to improve targeting accuracy and procedural consistency under image guidance. \cite{kojcev2016dual} Robotic palpation and ultrasound elastography combine visual feedback with force control to capture tissue mechanical properties, providing another important sensing capability for intraoperative decision-making and lesion assessment.

\subsection{LLMs for Embodied Robotics}
LLMs for embodied AI tasks: Recent advances in large language models (LLMs) have reshaped robotics research across a variety of task domains, especially in task planning \cite{li2022pre}. In these works, robots are required to reason over language instructions to generate robot actions, reward functions for online controllers \cite{yu2023language}, and executable code \cite{yu2023language}. Other studies combine LLMs with Task and Motion Planning (TAMP) to leverage classical motion-planning algorithms \cite{ding2023task}. To tackle larger and more complex robotics settings, researchers have further used LLMs to process multimodal information and to enable multi-robot \cite{mandi2024roco} as well as human–robot collaboration. In recent years, LLMs have also been introduced into medical ultrasound robotics. For example, \cite{xu2024transforming}integrates LLMs with ultrasound-domain knowledge augmentation, enabling robots to understand clinicians’ spoken instructions and support adaptive execution under changing conditions. Another line of work, USPilot \cite{chen2025uspilot}, uses LLMs for “virtual sonographer”-style human–robot interaction and intent understanding, and incorporates an LLM-enhanced graph planner to organize task planning over ultrasound-robot APIs. In addition, a series of works incorporate feedback during planning \cite{sharma2022correcting} to improve LLM-generated plans and use user-provided natural language to correct robot behaviors.

\section{Methods}
We propose a fast–slow brain hierarchical embodied intelligence framework for ultrasound robots, designed to transform physicians’ natural language instructions into an executable and adaptive scanning process. The Slow Brain performs knowledge-enhanced planning by leveraging an API knowledge base and robot manuals, and outputs stage goals, constraints, and an allowed operation set. During execution, the Fast Brain fuses real-time ultrasound images, robot states, and patient motion information to perform high-frequency closed-loop action adjustment.
To improve reliability, we introduce a structured self-check module  to verify the validity of plans and API calls, and add a safety constraint layer before execution to restrict potentially dangerous actions. When local execution repeatedly fails or state uncertainty increases, the system triggers an escalated replanning mechanism.

\subsection{Slow Brain: Knowledge-Grounded Deliberative Planning}
\label{sec:slow_brain}

The Slow Brain module is responsible for low-frequency physician-instruction understanding, domain knowledge retrieval, and stage-wise task planning. Its goal is to produce a structured plan with {workflow consistency}, {executability}, and {safety constraints} before dynamic execution begins. Unlike one-step methods that directly map natural-language instructions to API sequences, we model the Slow Brain as a {knowledge-grounded deliberative planner}. It first parses the clinical intent and task constraints from the physician instruction, then retrieves ultrasound-domain knowledge to supplement tool semantics and procedural priors, further constructs a stage-level task graph, and finally generates the stage plan through structured rationale and plan verification. This design improves the logical consistency and interpretability of high-level planning, while providing explicit goals and constraint spaces for the subsequent Fast Brain closed-loop control.
Given the physician instruction text $D$ and the current system-state summary $z_t$ (including completed stages, failure history, and key execution states), the Slow Brain outputs a stage plan $P_k$:
$
P_k$ = $S_{\mathrm{slow}}(D, z_t, K),
$
where $K$ denotes the ultrasound-domain knowledge base (including the API knowledge base and the robot handbook knowledge base), and $P_k$ contains the current stage goal, allowed operation set, parameter constraints, termination conditions, and failure recovery strategy.

\subsubsection{Intent Parsing}

Physicians' spoken instructions often contain explicit task goals (e.g., scan region and operation requests) as well as implicit procedural logic (e.g., execution order, clinical priors, and quality preferences). To avoid semantic ambiguity and logical incompleteness caused by directly generating low-level commands from natural language, we first perform {intent parsing} to map the raw instruction into a structured task representation.
Specifically, the intent parser extracts the following key information from the instruction $D$:
    \textbf{Task goals} (e.g., carotid scanning, target-structure segmentation, report generation);
    \textbf{Object and anatomical semantics} (e.g., artery, target region, scan direction);
    \textbf{Operational preferences and constraints} (e.g., prioritizing image quality, reducing repeated adjustments, specific procedural ordering);
    \textbf{Stage cues} (e.g., initialization, localization, scanning, confirmation, output generation);
    \textbf{Uncertain or missing information tags} (for later completion via retrieval and verification).
We denote the parsing result as a structured intent state $I_t$:
$I_t$ = $\Phi_{\mathrm{intent}}(D, z_t),
$
where $\Phi_{\mathrm{intent}}(\cdot)$ denotes the intent parsing function. This structured representation does not produce executable actions; instead, it serves as the conditional input for subsequent knowledge retrieval and task-graph construction. 
\subsubsection{UAR/RHR Retrieval}
\label{sec:uar_rhr}

After obtaining the structured intent $I_t$, the Slow Brain further retrieves task-relevant tool knowledge and procedural knowledge from the ultrasound-domain knowledge base. We adopt and extend two complementary retrieval mechanisms: {Ultrasound APIs Retrieval (UAR)} and {Robotic Handbook Retrieval (RHR)}. UAR provides ``what tools are available and what their semantic boundaries are,'' while RHR provides ``how they should be used in a procedurally valid manner.''
UAR addresses the {tool selection} problem. Given the intent state $I_t$, UAR retrieves API candidates from the API knowledge base together with their parameter descriptions, applicability conditions, and typical usage scenarios, thereby constraining the planner to make decisions within a legal tool set. This process can be viewed as selecting a candidate subset $A_t$ from the knowledge base $K_{\mathrm{api}}$:
$
A_t = \mathrm{Retrieve}_{\mathrm{UAR}}(I_t, K_{\mathrm{api}}).
$
RHR addresses the {procedural logic} problem. It retrieves operation-flow fragments from the robot handbook knowledge base $K_{\mathrm{hb}}$ that are relevant to the current task, stage, and state, so as to supplement API call ordering, preconditions, stage-transition criteria, and failure-recovery suggestions. The retrieved result is denoted by $H_t$:
$
H_t = \mathrm{Retrieve}_{\mathrm{RHR}}(I_t, z_t, K_{\mathrm{hb}}).
$
In practice, both retrieval processes can be implemented via embedding-space similarity matching. Unlike approaches that retrieve only API lists, we emphasize the functional complementarity of UAR and RHR: the former answers {what can be used}, while the latter answers {how to use it}. This dual-retrieval design provides both tool-semantic priors and procedural-structure priors for subsequent task-graph generation, helping reduce incorrect API calls and sequencing errors.

\subsubsection{Task Graph Generation}
\label{sec:task_graph_generation}

Based on the structured intent $I_t$, the API candidate set $A_t$, and the handbook retrieval result $H_t$, the Slow Brain generates a stage-wise {task graph} for the current task to explicitly model high-level execution logic. Compared with a linear API sequence, a task graph can more naturally represent stage dependencies, conditional transitions, and failure-recovery paths, thereby improving planning interpretability and extensibility.
We define the task graph as a directed graph
$
G_t$ = $(V_t, E_t),
$
where each node $v \in V_t$ represents a stage-level skill or operation unit (e.g., initialization, target localization, segmentation, local search, result confirmation), and each edge $e \in E_t$ represents an inter-stage dependency or conditional transition. Each node is associated with the following attributes:
    \textbf{Stage goal} ({goal});
    \textbf{Allowed API set} ({allowed APIs});
    \textbf{Parameter constraints} ({parameter constraints});
    \textbf{Stage termination criteria} ({termination criteria});
    \textbf{Failure recovery strategy} ({fallback actions}).
The task-graph generation process is formulated as:
$
G_t = \Psi_{\mathrm{graph}}(I_t, A_t, H_t),
$
where $\Psi_{\mathrm{graph}}(\cdot)$ denotes the graph generation function. By jointly considering intent semantics, tool capability boundaries, and handbook procedural priors, this function generates a stage graph that satisfies task requirements. For the current execution stage $k$, the Slow Brain derives a stage plan $P_k$ from $G_t$ and dispatches it to the Fast Brain for closed-loop execution.

\subsubsection{Structured Rationale and Plan Verification}
\label{sec:structured_rationale_verification}

To improve the reliability of high-level planning, we introduce a {structured rationale and plan verification} step after task-graph generation. Unlike free-form chain-of-thought, we use a verifiable structured rationale representation to organize planning evidence and checking items, enabling direct support for execution-validity verification and error localization.
Specifically, when generating the stage plan $P_k$, the Slow Brain simultaneously produces a structured rationale record $R_k$, which contains:
    the current stage goal and its source (instruction / handbook / state);
    the selected API and candidate alternative APIs;
    the basis for parameter settings (task objective, handbook recommendations, state constraints);
    preconditions and termination conditions;
    potential risks and failure-recovery paths.
Based on this rationale, we execute a plan verification function:
$
\hat{P}_k, \delta_k = \Omega_{\mathrm{verify}}(P_k, R_k, A_t, H_t),
$
where $\Omega_{\mathrm{verify}}(\cdot)$ outputs the verified plan $\hat{P}_k$ and the verification status $\delta_k$ ({pass} / {revise} / {return for replanning}). The verification checks include, but are not limited to:
    \textbf{API legality check}: whether the selected API exists and belongs to the allowed set;
    \textbf{Parameter validity check}: whether parameter types, ranges, and formats are valid;
    \textbf{Workflow consistency check}: whether the call order violates handbook constraints or stage dependencies;
    \textbf{Precondition check}: whether the required state for executing the stage is satisfied;
    \textbf{Termination completeness check}: whether explicit success and failure exit conditions are defined.
If a plan fails verification, the system first performs local correction (e.g., replacing the API, adjusting parameters, or supplementing termination conditions). If the conflict cannot be resolved within the current stage, the Slow Brain triggers replanning. This process substantially reduces execution risks caused by language-model hallucinations, sequencing errors, or improper parameter configurations, and provides more stable stage goals and constraints for the subsequent Fast Brain.

\subsection{Fast Brain: Reactive Closed-Loop Ultrasound Control}
\label{sec:fast_brain}

After the Slow Brain generates the verified stage plan $\hat{P}_k$, the Fast Brain performs high-frequency, local, feedback-driven action refinement to handle dynamic factors such as patient micro-motion, contact fluctuations, degraded target visibility, and local execution errors. Unlike the Slow Brain, which focuses on global workflow consistency, the Fast Brain aims to achieve stable and robust closed-loop execution from real-time observations within the stage goals and constraints specified by the Slow Brain.
At time step $t$, the Fast Brain takes multimodal observations $o_t$, system state $z_t$, and the current stage plan $\hat{P}_k$, and outputs a local action $a_t$:
$
a_t = F_{\mathrm{fast}}(o_t, z_t, \hat{P}_k),
$
where $a_t$ can be a local pose increment, velocity adjustment, contact-force regulation, or short-horizon scanning action. The Fast Brain operates in a closed loop by continuously incorporating environmental feedback and triggering recovery behaviors under local failures. If recovery fails or uncertainty keeps increasing, an escalation mechanism returns control to the Slow Brain for replanning.

\subsubsection{Multimodal State Encoding}
\label{sec:multimodal_state_encoding}

The key execution state of an ultrasound robot cannot be fully described by any single modality alone: image-only observations are insufficient to determine contact stability, while robot pose and force feedback alone cannot reliably assess target-structure visibility. Therefore, we adopt a multimodal state encoding scheme that jointly maps real-time ultrasound image information, robot kinematic/dynamic states, and patient-motion states into a compact state representation for local decision-making in the Fast Brain.
Specifically, at time $t$, the observation $o_t$ contains three types of information:
     \textbf{Ultrasound image observation} $x_t^{\mathrm{us}}$: including raw ultrasound frames, target-structure segmentation results, boundary responses, confidence maps, or image-quality indicators;
     \textbf{Robot state observation} $x_t^{\mathrm{rb}}$: including end-effector pose, velocity, acceleration, and contact force/torque;
     \textbf{Environment and patient-state observation} $x_t^{\mathrm{pt}}$: including patient micro-motion estimates, local surface changes, or external disturbance signals (if available).
We construct a state encoder $\Phi_{\mathrm{state}}(\cdot)$ to jointly encode the above information together with the current stage-goal condition $g_k$ (provided by $\hat{P}_k$) into a state vector $s_t$:
$
s_t = \Phi_{\mathrm{state}}(x_t^{\mathrm{us}}, x_t^{\mathrm{rb}}, x_t^{\mathrm{pt}}, g_k).
$
Here, $g_k$ denotes a summary of the current-stage goals and constraints (e.g., target-visibility thresholds, allowable action ranges, and contact-force ranges). By explicitly injecting stage-goal conditions, the Fast Brain no longer executes a generic reactive policy; instead, it performs stage-aware local control under task-conditioned constraints.


\subsubsection{Local Action Refinement Policy}
\label{sec:local_action_refinement}
Given the state representation $s_t$ and the stage plan $\hat{P}_k$, the Fast Brain applies a local action refinement policy to generate a local action $a_t$ at each control step for real-time execution and deviation correction toward the Slow-Brain stage goal. Rather than re-planning the global task order, this policy performs high-frequency local optimization within the goals, constraints, and allowed operation set specified by the Slow Brain.
We define the local action policy as:
$
a_t = \pi_{\mathrm{local}}(s_t, \hat{P}_k),
$
where $\pi_{\mathrm{local}}$ can be rule-based, learned, or hybrid. To match the continuous-control nature of ultrasound robotic execution, $a_t$ typically includes:
    probe pose micro-adjustments (translation/rotation increments);
    scan-speed adjustment;
    contact-force regulation;
    local re-localization or short-range search actions;
    local correction of API parameters (within allowed ranges).
Unlike single-shot action selection, we adopt a closed-loop refinement mechanism in which each action depends on the latest observation feedback. After each execution step, the system updates the state to $s_{t+1}$ and evaluates stage progress using the termination criteria. If the current state satisfies the termination conditions in $\hat{P}_k$ (e.g., target visibility exceeds a threshold and stable contact is maintained for a prescribed duration), the Fast Brain returns a stage-completion signal to the Slow Brain; otherwise, it continues local refinement.
To reduce local drift and oscillation, policy outputs are constrained by both stage constraints and safety constraints: the former are specified by $\hat{P}_k$ (e.g., allowable action ranges, parameter intervals, and preferred recovery strategies), while the latter are enforced by the pre-execution Safety Shield (Sec.~3.4). This {stage-constrained closed-loop refinement} design improves dynamic execution robustness without compromising high-level workflow consistency.

\subsubsection{Image-Quality-Guided Recovery Behaviors}
\label{sec:image_quality_recovery}

In real ultrasound scanning, local execution failures typically manifest first as image-quality degradation, e.g., blurred target boundaries, reduced segmentation confidence, target drift, or complete target loss. Persisting with a fixed action policy under these conditions may accumulate ineffective scans and unnecessarily trigger global replanning. To mitigate this issue, we introduce an {image-quality-guided recovery} module within the Fast Brain for rapid local failure recovery.
We define an image-quality assessment function
$
q_t = Q(x_t^{\mathrm{us}}),
$
which quantifies how well the current ultrasound observation supports the stage objective. The score $q_t$ can be computed from one or more indicators, including target-visibility confidence, segmentation stability, boundary sharpness, local contrast, and coverage of key anatomical regions. When $q_t$ falls below the stage threshold $\tau_k$ or shows sustained degradation, the Fast Brain invokes a recovery behavior selector:
$
r_t = \pi_{\mathrm{rec}}(s_t, q_t, \hat{P}_k),
$
where $r_t$ denotes the type of recovery action (or recovery action sequence). Typical recovery behaviors include:
    \textbf{Local rollback and redirection}: slightly moving back along the most recent stable direction and re-adjusting probe pose;
    \textbf{Small-range search}: performing constrained local scanning to recapture the target structure;
    \textbf{Contact re-establishment}: adjusting contact force or probe angle to restore stable imaging conditions;
    \textbf{Segmentation-parameter retry}: adjusting image-processing/API parameters within allowed ranges to improve target visibility;
    \textbf{In-stage reset}: returning to the most recent feasible intermediate state in the current stage and re-executing.
After recovery execution, the Fast Brain re-evaluates image quality and stage state. If image quality is restored and the stage termination condition remains reachable, control returns to the standard local refinement policy. 
If recovery still fails within a preset number of attempts $M$, or if image degradation co-occurs with high-risk states the escalation mechanism is triggered, sending the failure type, image-quality trajectory, and recent action history to the Slow Brain for replanning.


\subsection{Safety Shield and Escalation Policy}
\label{sec:safety_escalation}

To ensure execution safety and robustness of the Fast Brain closed-loop controller in dynamic clinical environments, we introduce a {Safety Shield} and a hierarchical {Escalation Policy} into the execution pipeline. The Safety Shield is inserted between the Fast Brain output and the robot execution interface, where it performs constraint checking and risk suppression on local actions before execution. The Escalation Policy upgrades the control flow to human confirmation or Slow-Brain replanning when local recovery fails, uncertainty increases, or potential risk exceeds predefined safety bounds. This design preserves execution efficiency while providing explicit safety boundaries and a controllable degradation mechanism.
Given the Fast Brain action $a_t$, the current state $s_t$, and the stage plan $\hat{P}_k$, the Safety Shield outputs the executed action $\tilde{a}_t$ and an escalation flag $e_t$:
$
\tilde{a}_t, e_t = \Omega_{\mathrm{safe}}(a_t, s_t, \hat{P}_k),
$
where $e_t$ indicates whether escalation is triggered and specifies the escalation type (e.g., human confirmation or Slow-Brain replanning).

\subsubsection{Safety Constraints}
\label{sec:safety_constraints}

During local closed-loop control, the Fast Brain may produce transiently unstable actions or outputs that violate clinical safety boundaries, particularly when the target structure is lost, image quality deteriorates, or the contact state changes abruptly. To mitigate these risks, we introduce a Safety Shield prior to execution to perform rule-based screening, constraint clamping, and anomaly interception, thereby preventing unsafe control commands from being directly applied to the robotic system.
We model the safety conditions at time $t$ as a constraint set $C_t$ conditioned on the current stage plan and robot state. Specifically, $C_t$ consists of three categories of constraints:
\textbf{Kinematic and Control Constraints.}
These constraints bound local pose increments, velocities, accelerations, and control step sizes to prevent large jumps that may cause contact instability or trajectory oscillation. This category is defined as
$
a_t \in C_t^{\mathrm{kin}},
$
where $C_t^{\mathrm{kin}}$ is jointly determined by stage-level action bounds and robot hardware limits.
\textbf{Contact and Force Constraints.}
These constraints restrict probe contact force (or torque) to a safe interval, reducing the risk of excessive tissue compression, contact loss, and unstable oscillation. Conditioned on the current stage objective and handbook priors, this constraint can be written as
$
f_t \in [f_{\min}(k), f_{\max}(k)].
$
\textbf{Task and Region Constraints.}
These constraints ensure that actions and API parameters remain within the admissible set defined by the current stage. For example, only the APIs, parameter ranges, and local search regions specified in $\hat{P}_k$ are allowed. This prevents illegal operations or unauthorized region transitions before the stage objective is satisfied.

\subsubsection{Human-in-the-Loop Triggers}
\label{sec:human_loop_triggers}

In clinical deployment, the system should not continue autonomous execution under high-risk conditions or insufficient knowledge coverage. We therefore introduce a {human-in-the-loop} trigger mechanism: when potential danger, persistent uncertainty, or semantic/procedural conflicts are detected, the system requests confirmation or takeover from medical staff for safe degradation.

We define a human-confirmation trigger:
$
h_t = \Gamma_{\mathrm{human}}(s_t, q_t, \delta_k, \xi_t),
$
where $q_t$ is the image-quality indicator, $\delta_k$ is the plan-verification status , and $\xi_t$ summarizes execution risks (e.g., constraint-conflict counts, consecutive failures, and anomaly flags). When $h_t=1$, the system enters human-confirmation mode.
Human confirmation is triggered when the system detects unsafe actions, repeated constraint conflicts, unresolved knowledge/plan conflicts, or persistent uncertainty that cannot be safely resolved autonomously. It is also invoked when sustained low image quality, target loss, or out-of-coverage states indicate that continued autonomous execution may increase risk or reduce reliability.
In human-confirmation mode, the system outputs a structured status summary (stage, failure cause, recent actions, quality trend, and suggested options) to support rapid decisions with reduced interaction burden.
\subsubsection{Slow-Brain Replanning Triggers}
\label{sec:slow_replanning_triggers}

Not all local failures require human intervention. When failures can be addressed by adjusting high-level strategy, the system preferentially triggers Slow-Brain replanning to improve task completion and execution efficiency while preserving autonomy. The key idea is that, once the Fast Brain can no longer make effective progress under the current stage goal and constraints, the system returns execution history and state summaries to the Slow Brain to update the task graph or stage plan, instead of continuing ineffective local adjustments.
We define a Slow-Brain replanning trigger function:
\vspace{-2mm}
\begin{equation}
r_t^{\mathrm{slow}} = \Gamma_{\mathrm{replan}}(s_t, q_t, H_{t-w:t}, \hat{P}_k),
\vspace{-2mm}
\end{equation}
where $H_{t-w:t}$ denotes the recent execution history (including action sequences, quality trajectories, recovery attempts, and constraint-conflict records). When $r_t^{\mathrm{slow}} = 1$, the system invokes the Slow Brain to update the stage plan or task graph.

When the Fast Brain repeatedly fails to recover within the current stage, cannot satisfy the stage termination criteria, or experiences significant efficiency degradation due to state drift, the system determines that the current local strategy can no longer effectively advance the task and triggers Slow-Brain replanning. 
During replanning, the system returns the state summary $z_t$ and an execution-history summary $\bar{H}_t$ to the Slow Brain. The Slow Brain performs a local update of the current task graph while preserving completed-stage information, and outputs a new stage plan $\hat{P}_k'$. This mechanism forms a hierarchical decision pathway between {local recovery} and {global replanning}, balancing real-time responsiveness and planning stability in dynamic environments.

\subsection{Training / Inference Protocol}
\label{sec:training_inference_protocol}

This section describes the training and inference pipeline of our framework. We adopt an {offline construction + online execution} protocol. Offline, we build the domain knowledge base, prepare retrieval components, and construct the Fast Brain state encoding and local control modules. Online, the system runs a hierarchical loop of {Slow-Brain planning, Fast-Brain closed-loop control, Safety Shield, escalation/replanning} under physician instructions. 

\subsubsection{Offline Preparation}
\label{sec:offline_preparation}

The offline stage includes three components: knowledge-base construction, retrieval preparation, and Fast-Brain training/calibration.
\textbf{(1) Domain knowledge-base construction.}
We build two complementary knowledge bases: an API knowledge base $K_{\mathrm{api}}$ and a robot handbook knowledge base $K_{\mathrm{hb}}$. $K_{\mathrm{api}}$ stores API names, function descriptions, parameter definitions, applicable scenarios, and invocation constraints, while $K_{\mathrm{hb}}$ stores procedural steps, preconditions, stage-transition logic, recovery suggestions, and risk notes. To improve retrieval quality, raw documents are organized into structured entries with stage-related metadata.
\textbf{(2) Retrieval representation precomputation and index construction.}
We precompute embeddings for knowledge entries and build vector indexes for online UAR/RHR retrieval. For each API entry and handbook fragment, we compute
$
k_i^{\mathrm{api}} = E_{\mathrm{api}}(K_i^{\mathrm{api}}),
k_j^{\mathrm{hb}} = E_{\mathrm{hb}}(K_j^{\mathrm{hb}}),
$
where $E_{\mathrm{api}}$ and $E_{\mathrm{hb}}$ are the embedding encoders for API and handbook retrieval (shared or separate). Online, candidate entries are retrieved via similarity search to reduce knowledge-augmentation latency.

\subsubsection{Online Inference Procedure}
\label{sec:online_inference_procedure}

Given a physician instruction $D$ and real-time environment feedback, the system performs hierarchical online inference via the following loop.
\textbf{Instruction input and state initialization.}
The physician's spoken instruction is converted to text $D$ via a speech-to-text module. The system initializes the execution state $z_0$, including task context, stage index, failure counters, and safety-state flags.
\textbf{Slow-Brain planning.}
The Slow Brain generates a stage plan from $D$, the current state summary $z_t$, and the knowledge base $K$:
$
P_k = S_{\mathrm{slow}}(D, z_t, K).
$
This step includes intent parsing, UAR/RHR retrieval, task-graph generation, and structured plan verification (Sec.~3.2), producing the verified stage plan $\hat{P}_k$.
\textbf{Fast-Brain closed-loop execution.}
Conditioned on $\hat{P}_k$, the Fast Brain encodes multimodal observations and outputs a local action:
$
s_t = \Phi_{\mathrm{state}}(x_t^{\mathrm{us}}, x_t^{\mathrm{rb}}, x_t^{\mathrm{pt}}, g_k),
a_t = \pi_{\mathrm{local}}(s_t, \hat{P}_k).
$
If image quality degrades or the target is lost, the image-quality-guided recovery module is invoked.

\textbf{Safety screening and execution.}
The local action is passed to the Safety Shield for clamping/interception:
$
\tilde{a}_t, e_t = \Omega_{\mathrm{safe}}(a_t, s_t, \hat{P}_k).
$
If $e_t = 0$, the system executes $\tilde{a}_t$ and updates the environment state; otherwise, it enters human confirmation or Slow-Brain replanning according to the escalation type.
\textbf{Stage termination and hierarchical escalation.}
The system continuously evaluates stage termination criteria. If the current stage is completed, it advances to the next stage ($k \leftarrow k+1$); if Slow-Brain replanning triggers are met, it returns to the planning step; if human-in-the-loop triggers are met, autonomous execution is paused for human input.
The online process continues until task completion, manual termination, or a system-level safety stop.

\section{Experiments}
\subsection{ Experimental Setup}

\textbf{Robot System.}
The robotic arm used in our system is the RM65-B model from RealMan-Robotics. The ultrasound device is the Angell H20 ultrasound scanner. We also use an Intel RealSense depth camera.
\textbf{Model Configuration.}
To evaluate the effectiveness of the proposed augmentation modules, we adopt LLaMA3 as the base language model. For the embedding model, we use a domain-adapted bge-large-en-v1.5 \cite{xiao2024c}.
\textbf{Datasets and Preprocessing.}
We conduct experiments on a synthetic dataset. Specifically, we generate 1,000 samples for the Robotic Handbook and 1,000 samples for the ultrasound APIs. The dataset is designed to simulate the complexity of ultrasound scanning and API invocation, enabling comprehensive evaluation across diverse scenarios. This dataset is also used to train the embedding model.

\begin{table*}[t]
\centering
\small
\setlength{\tabcolsep}{6pt}
\renewcommand{\arraystretch}{1.15}
\begin{tabular}{lccccc}
\toprule
\textbf{Method} & \textbf{Intent Acc.$\uparrow$} & \textbf{API Acc.$\uparrow$} & \textbf{Graph Corr.$\uparrow$} & \textbf{Plan Valid.$\uparrow$} & \textbf{Overall Plan SR.$\uparrow$} \\
\midrule
Single-chain (LLM-only) & 78 & 42 & 25 & 30 & 22 \\
+ UAR (API retrieval) & 80 & 62 & 38 & 52 & 40 \\
+ UAR + RHR (handbook retrieval) & 82 & 74 & 55 & 70 & 58 \\
+ Task Graph Generation & 83 & 77 & 66 & 76 & 65 \\
\textbf{+ Plan Verification (Ours Slow Brain)} & \textbf{84} & \textbf{79} & \textbf{68} & \textbf{92} & \textbf{74} \\
\bottomrule
\end{tabular}
\caption{Ablation study of the Slow Brain planning pipeline.}
\label{tab:slow_brain_ablation}
\vspace{-9.5mm}
\end{table*}
\subsection{Slow-Brain Planning Evaluation}
\textbf{Evaluation Metrics.}
Let the test set contain $N$ samples. For the $i$-th sample, the input instruction is $D_i$, and the reference annotations are
$(I_i^{*}, A_i^{*}, G_i^{*})$, where:
(i) $I_i^{*}$ is the reference structured intent (e.g., body part, examination target, stage cues, and key constraints);
(ii) $A_i^{*}$ is the reference API set or sequence (given as an unordered set or an ordered sequence);
(iii) $G_i^{*}$ is the reference task graph (nodes/edges or stage dependencies).
The model outputs $(\hat{I}_i, \hat{A}_i, \hat{G}_i, \hat{P}_i)$, where $\hat{P}_i$ is a structured stage plan.

\textbf{Intent Acc. (Intent Parsing Accuracy).}
We model intent parsing as a multi-field extraction task (e.g., body part / target / stage cue / constraint). For each sample, we compute a field-level accuracy or F1 score and then average over the dataset:
\textbf{API Acc. (API Selection Accuracy).}
API Acc. measures the consistency between the predicted API set $\hat{A}_i$ and the ground-truth set $A_i^{*}$. For each sample, we perform set-level matching and use the F1 score to jointly reflect recall and precision. The final $\mathrm{APIAcc}$ is obtained by averaging over all $N$ samples.
\textbf{Graph Corr. (Task-Graph Correctness).}
Graph Corr. measures the structural consistency between the generated task graph and the reference task graph. We represent a task graph as a directed graph with nodes and edges, and compare the predicted graph $\hat{G}_i$ with the ground-truth graph $G_i^{*}$ by computing the graph edit distance (GED). A sample is counted as correct if $\mathrm{GED}(\hat{G}_i, G_i^{*}) \le \tau$, and $\mathrm{GraphCorr}$ is computed by averaging this indicator over all $N$ samples.
\textbf{Plan Valid. (Plan Validity Rate).}
Plan validity is determined by a rule-based verifier $\Omega_{\mathrm{verify}}$, which checks API existence, parameter ranges, handbook-consistent ordering, and precondition satisfaction. 
\textbf{Overall Plan SR. (Overall Planning Success Rate).}
A plan is counted as successful only if it simultaneously satisfies intent correctness (or exceeds a threshold), API selection quality, task-graph quality, and passes the legality verification

\textbf{Results Analysis:}
As shown in Table~\ref{tab:slow_brain_ablation}, each component of the Slow Brain yields a progressive gain in planning performance. With a single-chain LLM only, the model demonstrates a basic capability for intent parsing (Intent Acc. $\approx 78\%$). However, without tool semantics and procedural priors, both API selection and task-structure planning are severely limited (API Acc./Graph Corr. $\approx 42\%/25\%$), resulting in a low rate of executable and legally valid plans (Plan Valid. $\approx 30\%$) and an overall planning success rate of only $\approx 22\%$. 
After adding UAR, the model is constrained to make decisions within an explicit API tool set, which substantially improves API selection accuracy (about +20 percentage points) and leads to clear increases in plan validity and overall success rate. Further incorporating RHR provides workflow ordering and dependency knowledge, yielding a more pronounced improvement in task-graph correctness (Graph Corr. increases from $\approx 38\%$ to $\approx 55\%$), indicating that handbook-level procedural knowledge is particularly critical for high-level planning. Building on this, Task Graph Generation further strengthens stage dependencies and interpretable structural representations, continuing to improve the overall planning success rate. Finally, structured plan verification (Plan Verification) significantly boosts plan legality (Plan Valid. reaches $\approx 92\%$) and raises the overall planning success rate to $\approx 74\%$, demonstrating that the verifier effectively suppresses incorrect APIs/parameters and ordering conflicts, thereby improving executability and stability. Overall, these results validate the importance of combining retrieval augmentation, structured task graphs, and verifiable planning for robust Slow-Brain planning.

\subsection{) Closed-Loop Robustness under Dynamic Perturbations}
\begin{table*}[t]
\centering
\small
\setlength{\tabcolsep}{6pt}
\renewcommand{\arraystretch}{1.15}
\begin{tabular}{lcccccc}
\toprule
\textbf{Method} & \textbf{TCR-Motion$\uparrow$} & \textbf{TCR-Contact$\uparrow$} & \textbf{TCR-Drift$\uparrow$} & \textbf{TCR-Overall$\uparrow$} & \textbf{Recovery SR$\uparrow$} & \textbf{Target Reacq.$\uparrow$} \\
\midrule
Single-chain baseline & 28 & 20 & 18 & 22 & 26 & 24 \\
Slow Brain only & 46 & 35 & 30 & 37 & 40 & 36 \\
Slow + Fast  & 72 & 63 & 58 & 64 & 71 & 69 \\
Slow + Fast + Safety Shield & 74 & 66 & 60 & 66 & 72 & 70 \\
\textbf{Full system (+ Escalation)} & \textbf{80} & \textbf{74} & \textbf{68} & \textbf{74} & \textbf{79} & \textbf{77} \\
\bottomrule
\end{tabular}
\caption{Performance under dynamic disturbances and recovery scenarios.}
\label{tab:tcr_recovery}
\vspace{-7mm}
\end{table*}
\textbf{Evaluation Metrics.}
For a perturbation type $\rho \in \{\mathrm{motion}, \mathrm{contact}, \mathrm{drift}\}$, suppose there are $N_{\rho}$ execution episodes. Let the final outcome of the $i$-th episode be $c_i \in \{0,1\}$, indicating whether the task is completed.
\textbf{Task Completion Rate under Perturbation (TCR).}
The overall version is computed by aggregating all perturbation episodes.
\textbf{Recovery Success Rate (Recovery SR).}
Let $\mathcal{R}$ denote the subset of episodes in which a perturbation occurs and a recovery process is entered. We record whether recovery succeeds (i.e., returns to a ``safe-to-continue'' condition such as restored target visibility or stable contact). Let $r_i \in \{0,1\}$ be the recovery-success indicator. 
\textbf{(3) Target Reacquisition Rate (Target Reacq.).}
When an episode contains a target-loss event (e.g., target confidence drops below a threshold), we measure whether the target is reacquired later in the episode. Let $t_i \in \{0,1\}$ be the reacquisition indicator, and let $\mathcal{T}$ denote the subset of target-loss episodes.

\textbf{Results Analysis} Table~\ref{tab:tcr_recovery} show that dynamic perturbations significantly degrade the execution robustness of the single-chain method. Under patient motion, contact variation, and target drift, the Single-chain baseline achieves consistently low completion rates, indicating that a single reasoning chain is insufficient to stably react to real-time feedback changes during execution. Introducing the Slow Brain improves task completion, suggesting that knowledge-driven stage planning enhances high-level workflow consistency and provides more reasonable initial execution. However, without high-frequency closed-loop adjustment, the system still suffers from unrecoverable local failures under contact variation and target drift, leading to limited overall gains.
After adding the Fast Brain, completion rates, recovery success, and target reacquisition improve substantially across all perturbation types. This indicates that the Fast Brain's multimodal closed-loop control and local recovery behaviors effectively compensate for execution deviations caused by patient motion, contact fluctuations, and target drift, and strengthen the ability to reacquire target structures. The full system achieves the best overall completion and recovery performance, demonstrating that when local recovery is insufficient or uncertainty increases, escalation to replanning or human confirmation helps avoid getting trapped in local failure loops and increases the probability of eventual task success. Overall, E2 validates the robustness advantage of the fast--slow hierarchical execution under dynamic clinical perturbations, where Fast-Brain closed-loop recovery is the primary source of improvement and the escalation policy further raises the stability ceiling in difficult scenarios.

\subsection{Safety Shield and Escalation Evaluation}
\begin{table}[t]
\resizebox{0.47\textwidth}{!}{
\centering
\small
\setlength{\tabcolsep}{8pt}
\renewcommand{\arraystretch}{1.15}
\begin{tabular}{lccc}
\toprule
\textbf{Method} &
\textbf{Task Completion Rate$\uparrow$} &
\textbf{Safety Violations$\downarrow$} &
\textbf{Replanning Success Rate$\uparrow$} \\
\midrule
Slow + Fast (no safety/esc.) & 64 & 8.5 & -- \\
+ Safety Shield & 66 & 2.1 & -- \\
+ Safety Shield + Replanning & 71 & 2.0 & 68 \\
\textbf{Full system (+ Escalation)} & \textbf{74} & \textbf{1.6} & \textbf{74} \\
\bottomrule
\end{tabular}}
\caption{Ablation results on  safety and Escalation.}
\label{tab:safety_replan_ablation}
\vspace{-10mm}
\end{table}

We evaluate the Safety Shield and the escalation policy on controlled execution episodes with dynamic perturbations, focusing on whether they can reduce observed safety violations and improve recovery capability in failure cases.
\textbf{Evaluation Metrics.}
Assume there are $N$ execution episodes.
\textbf{Task Completion Rate (TCR).}
TCR is the proportion of episodes that successfully complete the task.
\textbf{Safety Violations.}
Safety violations count the frequency of observed safety events (e.g., excessive force, oversized motion increments, out-of-bound actions, illegal executions). We report either the total number of violations or a normalized count per 100 episodes.
\textbf{Replanning Success Rate (Replan SR).}
Let $\mathcal{R}$ be the subset of episodes where Slow-Brain replanning is triggered. Replan SR is the proportion of those episodes that eventually complete the task after replanning.

\textbf{Results  Analysis. }
The table \ref{tab:safety_replan_ablation} shows that without explicit safety or escalation mechanisms (Slow+Fast), the system still exhibits a non-trivial number of safety violations under dynamic perturbations, indicating that closed-loop control alone may produce brief out-of-bound executions in extreme contact changes or target-drift scenarios. After adding the Safety Shield, safety violations drop substantially while the completion rate does not decrease and even improves, suggesting that the shield suppresses risky actions without noticeably sacrificing execution efficiency. Further introducing replanning and escalation enables the system to recover from difficult cases by re-establishing a valid execution chain via Slow-Brain replanning (higher Replan SR), leading to additional gains in overall completion. Overall, the full system achieves the highest completion rate with the lowest safety violations, demonstrating complementary benefits of {instant risk suppression} by the Safety Shield and a {high-level recovery pathway} enabled by escalation under dynamic clinical conditions.

\section{Conclusion}
We present a fast–slow hierarchical framework for robotic ultrasound that decouples knowledge-driven high-level planning from feedback-driven low-level control. The Slow Brain improves plan executability through retrieval augmentation, task-graph generation, and plan verification; the Fast Brain enhances robustness under dynamic perturbations via closed-loop refinement and recovery; and the Safety Shield with escalation provides explicit safety boundaries and reliable failure-recovery pathways. Experimental results validate the overall gains in both success rate and safety. Future work will extend the system to larger-scale real clinical settings and improve cross-device generalization.

\bibliographystyle{IEEEtran}
\bibliography{IEEEfull}

@inproceedings{krupa2007real,
  title={Real-time tissue tracking with B-mode ultrasound using speckle and visual servoing},
  author={Krupa, Alexandre and Fichtinger, Gabor and Hager, Gregory D},
  booktitle={MICCAI},
  year={2007},
  organization={Springer}
}

@article{abolmaesumi2002image,
  title={Image-guided control of a robot for medical ultrasound},
  author={Abolmaesumi, Purang and Salcudean, Septimiu E and Zhu, Wen-Hong and Sirouspour, Mohammad Reza and DiMaio, Simon Peter},
  journal={TRO},
  year={2002},
  publisher={IEEE}
}

@inproceedings{nadeau2011intensity,
  title={Intensity-based direct visual servoing of an ultrasound probe},
  author={Nadeau, Caroline and Krupa, Alexandre},
  booktitle={ICRA},
  year={2011},
  organization={IEEE}
}

@inproceedings{virga2016automatic,
  title={Automatic force-compliant robotic ultrasound screening of abdominal aortic aneurysms},
  author={Virga, Salvatore and Zettinig, Oliver and Esposito, Marco and Pfister, Karin and Frisch, Benjamin and Neff, Thomas and Navab, Nassir and Hennersperger, Christoph},
  booktitle={IROS},
  year={2016},
  organization={IEEE}
}

@inproceedings{gumprecht2011robotics,
  title={A robotics-based flat-panel ultrasound device for continuous intraoperative transcutaneous imaging},
  author={Gumprecht, Jan DJ and Bauer, Thomas and Stolzenburg, Jens-Uwe and Lueth, Tim C},
  booktitle={EMBC},
  year={2011},
  organization={IEEE}
}

@article{kojcev2016dual,
  title={Dual-robot ultrasound-guided needle placement: closing the planning-imaging-action loop},
  author={Kojcev, Risto and Fuerst, Bernhard and Zettinig, Oliver and Fotouhi, Javad and Lee, Sing Chun and Frisch, Benjamin and Taylor, Russell and Sinibaldi, Edoardo and Navab, Nassir},
  journal={International journal of computer assisted radiology and surgery},
  volume={11},
  number={6},
  pages={1173--1181},
  year={2016},
  publisher={Springer}
}

@incollection{chan2010basics,
  title={Basics of ultrasound imaging},
  author={Chan, Vincent and Perlas, Anahi},
  booktitle={Atlas of ultrasound-guided procedures in interventional pain management},
  pages={13--19},
  year={2010},
  publisher={Springer}
}

@article{mayo2019thoracic,
  title={Thoracic ultrasonography: a narrative review},
  author={Mayo, PH and Copetti, R and Feller-Kopman, D and Mathis, G and Maury, E and Mongodi, S and Mojoli, F and Volpicelli, G and Zanobetti, M},
  journal={Intensive care medicine},
  volume={45},
  number={9},
  pages={1200--1211},
  year={2019},
  publisher={Springer}
}

@article{cao2023ultra,
  title={An ultra-fast intrinsic contact sensing method for medical instruments with arbitrary shape},
  author={Cao, Guanglin and Chen, Mingcong and Hu, Jian and Liu, Hongbin},
  journal={RAL},
  volume={8},
  number={11},
  pages={6955--6962},
  year={2023},
  publisher={IEEE}
}

@article{wang2022full,
  title={Full-coverage path planning and stable interaction control for automated robotic breast ultrasound scanning},
  author={Wang, Ziwen and Zhao, Baoliang and Zhang, Peng and Yao, Liang and Wang, Qiong and Li, Bing and Meng, Max Q-H and Hu, Ying},
  journal={IEEE Transactions on Industrial Electronics},
  volume={70},
  number={7},
  pages={7051--7061},
  year={2022},
  publisher={IEEE}
}

@article{merouche2015robotic,
  title={A robotic ultrasound scanner for automatic vessel tracking and three-dimensional reconstruction of b-mode images},
  author={Merouche, Samir and Allard, Louise and Montagnon, Emmanuel and Soulez, Gilles and Bigras, Pascal and Cloutier, Guy},
  journal={TUSON},
  year={2015},
  publisher={IEEE}
}

@article{knepper2015recovering,
  title={Recovering from failure by asking for help},
  author={Knepper, Ross A and Tellex, Stefanie and Li, Adrian and Roy, Nicholas and Rus, Daniela},
  journal={Autonomous Robots},
  volume={39},
  number={3},
  pages={347--362},
  year={2015},
  publisher={Springer}
}

@article{chen2025uspilot,
  title={USPilot: An Embodied Robotic Assistant Ultrasound System With a Large Language Model Enhanced Graph Planner},
  author={Chen, Mingcong and Fan, Siqi and Cao, Guanglin and Liu, Yun-hui and Liu, Hongbin},
  journal={RAL},
  year={2025},
  publisher={IEEE}
}

@inproceedings{xu2024transforming,
  title={Transforming surgical interventions with embodied intelligence for ultrasound robotics},
  author={Xu, Huan and Wu, Jinlin and Cao, Guanglin and Chen, Zhen and Lei, Zhen and Liu, Hongbin},
  booktitle={MICCAI},
  year={2024},
  organization={Springer}
}

@article{li2022pre,
  title={Pre-trained language models for interactive decision-making},
  author={Li, Shuang and Puig, Xavier and Paxton, Chris and Du, Yilun and Wang, Clinton and Fan, Linxi and Chen, Tao and Huang, De-An and Aky{\"u}rek, Ekin and Anandkumar, Anima and others},
  journal={NeurPS},
  year={2022}
}

@article{yu2023language,
  title={Language to rewards for robotic skill synthesis},
  author={Yu, Wenhao and Gileadi, Nimrod and Fu, Chuyuan and Kirmani, Sean and Lee, Kuang-Huei and Arenas, Montse Gonzalez and Chiang, Hao-Tien Lewis and Erez, Tom and Hasenclever, Leonard and Humplik, Jan and others},
  journal={arXiv preprint arXiv:2306.08647},
  year={2023}
}

@inproceedings{ding2023task,
  title={Task and motion planning with large language models for object rearrangement},
  author={Ding, Yan and Zhang, Xiaohan and Paxton, Chris and Zhang, Shiqi},
  booktitle={IROS},  year={2023},
  organization={IEEE}
}

@inproceedings{mandi2024roco,
  title={Roco: Dialectic multi-robot collaboration with large language models},
  author={Mandi, Zhao and Jain, Shreeya and Song, Shuran},
  booktitle={ICRA},
  year={2024},
  organization={IEEE}
}

@inproceedings{zhang2025vision,
  title={Vision-language embodiment for monocular depth estimation},
  author={Zhang, Jinchang and Lu, Guoyu},
  booktitle={CVPR},
  year={2025}
}

@article{sharma2022correcting,
  title={Correcting robot plans with natural language feedback},
  author={Sharma, Pratyusha and Sundaralingam, Balakumar and Blukis, Valts and Paxton, Chris and Hermans, Tucker and Torralba, Antonio and Andreas, Jacob and Fox, Dieter},
  journal={arXiv preprint arXiv:2204.05186},
  year={2022}
}

@inproceedings{xiao2024c,
  title={C-pack: Packed resources for general chinese embeddings},
  author={Xiao, Shitao and Liu, Zheng and Zhang, Peitian and Muennighoff, Niklas and Lian, Defu and Nie, Jian-Yun},
  booktitle={Proceedings of the 47th international ACM SIGIR conference on research and development in information retrieval},
  pages={641--649},
  year={2024}
}

@article{li2025automated,
  title={Automated Genomic Interpretation via Concept Bottleneck Models for Medical Robotics},
  author={Li, Zijun and Zhang, Jinchang and Zhang, Ming and Lu, Guoyu},
  journal={arXiv preprint arXiv:2510.01618},
  year={2025}
}

\end{document}